%% file: root.tex
\documentclass[letterpaper, 10 pt, conference]{ieeeconf}  

\IEEEoverridecommandlockouts                              

\usepackage[utf8]{inputenc}

\usepackage{graphics} 
\usepackage{epsfig} 
\usepackage{times} 
\usepackage{amsmath} 
\usepackage{amssymb}  
\usepackage{multirow}
\usepackage{diagbox}
\usepackage{svg}
\usepackage{caption}
\usepackage{tikz}
\usepackage{subcaption}
\usepackage{float}
\usepackage{wrapfig}

\usepackage{cite}
\usepackage[dvipsnames]{xcolor}
\usepackage{soul}
\usepackage[noend]{algpseudocode}
\usepackage[ruled,vlined,linesnumbered]{algorithm2e}
\usepackage{algpseudocode}

\usepackage{color}
\usepackage{tikz}
\usepackage{comment}
\usepackage{hyperref}
\usepackage{booktabs}
\usepackage{adjustbox}
\usepackage{gensymb}
\usepackage{caption} 

\newcommand{\statevec}{\mathbf{x}}
\newcommand{\ourmethodto}{UGE-TO}
\newcommand{\ourmethodmpc}{UGE-MPC}
\newcommand{\ourmethodlong}{Uncertainty Guided Exploratory Trajectory Optimization}

\title{\LARGE \bf
Uncertainty Guided Exploratory Trajectory Optimization for Sampling-Based Model Predictive Control
}

\author{O. Goktug Poyrazoglu$^{1}$,  Yukang Cao$^{1}$, Rahul Moorthy$^{1}$ and Volkan Isler$^{2}$
\thanks{The authors are with the Robotics, Sensing and Networks Laboratory (RSN).
        $^{1}$University of Minnesota, Minneapolis, MN 55455, USA.
        $^{2}$The University of Texas at Austin, Austin, TX 78712, USA.
        Correspondence: {\tt\small poyra002@umn.edu}.}%
}

\begin{document}

\maketitle
\thispagestyle{empty}
\pagestyle{empty}

\begin{abstract}
\input{sections/abstract}
\end{abstract}

\vspace{-3pt}
\section{Introduction}
\input{sections/introduction}
\section{Related Work}
\input{sections/relatedwork}

\section{Problem Formulation}
\input{sections/problem_formulation}
\section{Uncertainty Guided Exploratory Trajectory Optimization} \label{sec:uaeto}

\input{sections/approach}

\section{Uncertainty Guided Exploratory Model Predictive Control} \label{sec:uaempc}
\input{sections/uae_mpc}
\vspace{-2pt}
\section{EXPERIMENTS}
\input{sections/experiments}

\section{CONCLUSION}
\input{sections/conclusion}




\bibliographystyle{unsrt}
\bibliography{root}

\end{document}

%% file: sections/abstract.tex
Trajectory optimization depends heavily on initialization. In particular, sampling-based approaches are highly sensitive to initial solutions, and limited exploration frequently leads them to converge to local minima in complex environments. We present~\ourmethodlong{}~(\ourmethodto{}), a trajectory optimization algorithm 
that generates well-separated samples to achieve a better coverage of the configuration space.~\ourmethodto{}~represents trajectories as probability distributions induced by uncertainty ellipsoids. Unlike sampling-based approaches that explore only in the action space, this representation captures the effects of both system dynamics and action selection. By incorporating the impact of dynamics, in addition to the action space, into our distributions, our method enhances trajectory diversity by enforcing distributional separation via the Hellinger distance between them. It enables a systematic exploration of the configuration space and improves robustness against local minima. 
Further, we present~\ourmethodmpc{}, which integrates~\ourmethodto{} into sampling-based model predictive controller methods. Experiments demonstrate that~\ourmethodmpc{} achieves higher exploration and faster convergence in trajectory optimization compared to baselines under the same sampling budget, achieving 72.1\% faster convergence in obstacle-free environments and 66\% faster convergence with a 6.7\% higher success rate in the cluttered environment compared to the best-performing baseline.
Additionally, we validate the approach through a range of simulation scenarios and real-world experiments. Our results indicate that~\ourmethodmpc{} has higher success rates and faster convergence, especially in environments that demand significant deviations from nominal trajectories to avoid failures. 
The project and code are available at \url{https://ogpoyrazoglu.github.io/cuniform_sampling/}. 

%% file: sections/introduction.tex
Trajectory optimization is a powerful concept in robotics for generating efficient robot motions under task-specific objectives. it has been successfully applied to applications including navigation~\cite{chomp}, manipulation~\cite{dragan_manipulation_2011}, and aerial robotics~\cite{sanchez-lopez_trajectory_2020}. Classical gradient-based methods focus on smooth settings, where differentiability can be exploited for efficient optimization, and have demonstrated strong performance in such settings~\cite{schulman_motion_2014}-~\cite{gusto}. However, their reliance on differentiability limits their direct use in complex multimodal environments. In contrast, sampling-based approaches relax these assumptions by allowing non-differentiable cost functions, making them applicable to a broader range of robotic systems and tasks. They are most commonly applied in Model Predictive Control (MPC) settings since they can handle nonlinear dynamics with arbitrary cost functions~\cite{mppi}. 

Despite these advantages, sampling-based methods remain highly sensitive to the selection of the underlying sampling distribution and the initialization. 
For example, Model Predictive Path Integral (MPPI)~\cite{mppi} samples action sequences from a Gaussian distribution centered on the nominal trajectory, resulting in trajectories that are concentrated around it. As a result, when the distribution lacks sufficient diversity or the nominal trajectory is poorly initialized, the resulting trajectories may fail to cover promising areas of the configuration space. This often leads to convergence to local minima and reduces the controller's reliability in complex environments. Addressing this limitation is crucial to enhancing the robustness of sampling-based approaches.

\begin{figure}[t]
    \includegraphics[width=\linewidth, height=5.5 cm]{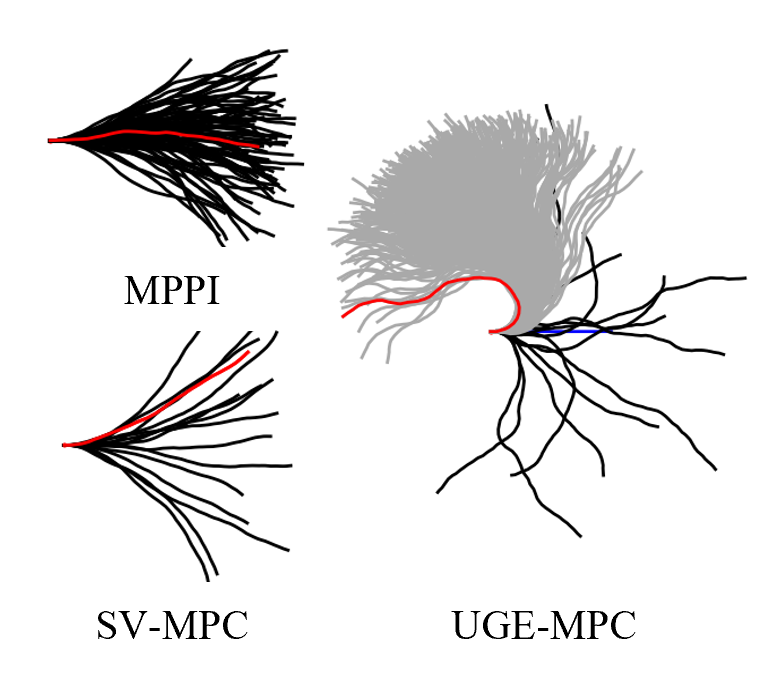}
    \caption{Comparison of first-iteration trajectories when the robot starts heading +x, and the nominal trajectory is forward, with the goal located behind the robot. Under the same sample budget of $2048$. Red trajectories indicate the solutions obtained in this iteration for each method. Black trajectories denote the sampled particles for SV-MPC and \ourmethodmpc{}, and the rollouts for MPPI. Leveraging distributional separation in trajectory space, ~\ourmethodmpc{} produces backward-reaching candidates and a feasible solution in one iteration. SV-MPC enhances diversity through repulsion but does not consistently cover the required full-turn maneuvers, whereas MPPI remains concentrated around the nominal trajectory.}
    \label{fig:figure_1}
    \vspace{-12pt}
\end{figure}

Several methods have been proposed in the literature to enhance exploration, thereby improving the reliability of the methods. Early approaches directly modify the sampling distribution to encourage broader exploration. For example, Log-MPPI~\cite{logmppi}, which increases trajectory diversity by reshaping the underlying distribution. Another line of work is to incorporate entropy-based regularization terms into the objective to promote diversity in the resulting policies~\cite{pmlr-v70-haarnoja17a, lambert2021entropyregularizedmotionplanning}, thereby penalizing narrow policy distributions. More recently, multimodal approaches have been presented to capture a distribution over trajectories rather than a single solution. A class of approaches~\cite{svmpc, svddp} uses Stein Variational Gradient Descent (SVGD)~\cite{svgd}, which uses particles to approximate multimodal trajectory distributions by the repulsive feature of SVGD to maintain the diversity among trajectories, while shifting them toward low-cost regions. While these methods improve exploration, they often rely solely on action-space perturbation and may not fully capture the impact of system dynamics on trajectory diversity.    

In this work, we present a new exploration technique that
extends the idea of exploration via action perturbations by
accounting for the impact of long-range system dynamics. Specifically, we introduce~\ourmethodlong{}~(\ourmethodto{}) that constructs probability distributions for trajectories by using uncertainty ellipsoids around trajectories. After the trajectory distributions are constructed, we enforce distributional separation using the Hellinger Distance to reduce overlap between the distributions. In other words, it will lead to exploration of the configuration space by forcing trajectories to visit distinct regions of the configuration space. We further present~\ourmethodmpc{}, which integrates~\ourmethodto{} into a sampling-based MPC framework. In this setting, we use~\ourmethodto{} to generate diverse trajectory candidates and identify a better nominal trajectory, which helps the controller to perform a local search around low-cost regions.

In summary, our contributions of this work are:
\begin{itemize}
    \item We introduce~\ourmethodlong{} (\ourmethodto{}), which utilizes the distribution separation of trajectory probability distributions generated by uncertainty ellipsoids to effectively explore the configuration space.
    \item We present~\ourmethodmpc{}, a new sampling-based model predictive controller that utilizes \ourmethodto{} to enhance exploration of representative trajectory rollouts and then performs a local search around the selected minimum cost one. By achieving high exploration of the configuration space, our method reduces the likelihood of becoming trapped in local minima.
    \item We evaluate our approach in simulation and real-world scenarios,
    demonstrating improved exploration, faster convergence, and higher success rates compared to baselines, particularly in environments requiring significant deviations from nominal trajectories. 
\end{itemize}

Integrating \ourmethodto{} into a sampling-based MPC (\ourmethodmpc{}) provides a systematic approach that augments exploration with local search to refine exploitation. We begin with an overview of the related work.

%% file: sections/relatedwork.tex

\subsection{Trajectory Optimization}

Existing approaches to local trajectory optimization can be divided into two branches: direct and indirect methods~\cite{to-general}. One approach is to use direct methods that translate the optimization problem into a nonlinear problem~\cite {schulman_motion_2014}, \cite{gusto}, and solve it with nonlinear solvers, such as IPOPT~\cite{ipopt} and SNOPT~\cite{snopt}. On the other side, indirect methods aim to convert the problem into subproblems and solve for the local optimality conditions. Differential dynamic programming (DDP)~\cite{DDP} and iterative Linear Quadratic Regulator (iLQR)~\cite{ilqr} are well-known examples of this method. A major limitation of both methods is their strong dependence on initialization. Since they seek to find a single local-optimal solution, this property increases the importance of the initialization. Later work combines maximum entropy policies with DDP and avoids local minima via exploration capacity of their multimodal policy~\cite{so2022maximumentropydifferentialdynamic},~\cite{aoyama2024generalizedmaximumentropydifferential}. Similarly, CSVTO~\cite{power2024constrainedsteinvariationaltrajectory} uses Stein Variational Gradient Descent (SVGD)~\cite{svgd} to generate a diverse set of trajectories to avoid possible local minima. Our work also aligns with this line of research by aiming to overcome the limitations of a single local solution. However, in contrast to these methods that rely on convexity or differentiability of the cost functions, we focus on sampling-based approaches that naturally handle non-convex and non-differentiable objectives. This allows us to enhance diversity and robustness in trajectory optimization without depending on restrictive assumptions. 

\subsection{Sampling-Based Trajectory Optimization}

Sampling-based approaches have gained popularity and become state-of-the-art for handling non-smooth cost landscapes by directly evaluating candidate control sequences drawn from their distributions~\cite{kazim2024recent}. Early methods, such as the Cross-Entropy Method (CEM)~\cite{cem} and Model Predictive Path Integral control (MPPI)~\cite{mppi}, demonstrated strong performance across various robotic systems by iteratively adapting their action-sampling distributions based on rollout costs. However, similar to gradient-based methods, these approaches also struggle with local minima in the cost landscape due to their limited exploration in the configuration space. Recent works have sought to address these issues by improving sampling strategies. One line of research focused on directly modifying the sampling distribution. Log-MPPI~\cite{logmppi} leverages the Normal-Log-Normal (NLN) distribution to increase the sample diversity. Similarly, colored-noise-based variants use correlated noises to generate both smoother and diverse trajectories~\cite{Vlahov_2024},~\cite{colorednoise}. Although these methods have increased the exploration capabilities, they still rely on unimodal sampling strategies that lead to a single solution. 

On the other hand, another class of methods introduces multimodal sampling distributions. In the SVGD-based approach, SV-MPC~\cite{svmpc} uses particle approximations of trajectories to represent a multimodal trajectory distribution with multiple particles. SV-MPC's MPPI extension, SVG-MPPI~\cite{honda2024steinvariationalguidedmodel}, guides trajectories to low-cost regions by modifying both the nominal trajectory and the covariances. Similarly, C-Uniform trajectory sampling~\cite{poyrazoglu2024cuniformtrajectorysamplingfast} has been presented to maintain uniform coverage of configurations over level sets, and that leads to high exploration over the configuration space. Its extension, CU-MPPI~\cite{poyrazoglu2025unsupervisedcuniformtrajectorysampler}, incorporates C-Uniform trajectories for better nominal trajectory selection into the MPPI framework to avoid local minima. In contrast, \ourmethodmpc{} applies Hellinger distance–based separation directly in the space of trajectory distributions, ensuring diversity under both actions and dynamics. This approach avoids the scalability limitations of entropy-based grid methods, such as CU-MPPI, while providing more principled configuration-space clustering than action-level kernels, like SV-MPC. Another method closely related to our work is U-MPPI~\cite{mohamed2024efficientmppitrajectorygeneration}, which utilizes the unscented transform to enhance state-space exploration. Compared with U-MPPI, our method also uses uncertainty ellipsoids, but unlike U-MPPI, we explicitly represent trajectories as probability distributions and separate them distributionally using the Hellinger Distance-based metric. This property explicitly prevents overlap between trajectory distributions, which promotes exploration and reduces the likelihood of collapsing into a local minimum.

%% file: sections/problem_formulation.tex
\begin{figure*}[t]
    \centering
    \includegraphics[width=\linewidth, height=4cm]{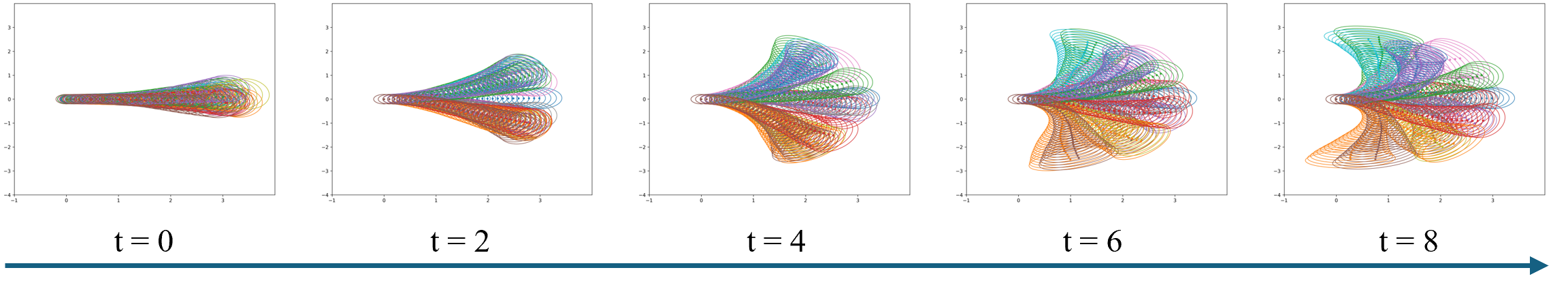}
    \caption{Illustration of the Hellinger Distance-based Distributional Separation: Candidate rollouts (leftmost, $t=0$) are represented as trajectory distributions with uncertainty ellipsoids. Optimization reduces overlap among distributions, resulting in a diverse set of nominal trajectories. The figure demonstrates how the distributional separation happens and presents two-dimensional projections of the three-dimensional state space $[x,y,\theta]$. The final trajectory distributions cover the configuration space more broadly, which enables better exploration and a nominal trajectory selection for SBMPCs.}
    \label{fig:method_figure}
\end{figure*}

We begin by formulating trajectory optimization as a general stochastic optimal control problem, where $\mathbf{x}_t \in \mathbb{R}^n$ denotes the system state and $\mathbf{u}_t \in \mathbb{R}^m$ denotes the control input at discrete time step $t$. The system evolves according to the following stochastic dynamics:

\begin{equation} \label{eq:system_dynamics_general}
    \mathbf{x}_{t+1} = f(\mathbf{x}_t, \mathbf{u}_t) + \boldsymbol{\epsilon}_t, \quad \boldsymbol{\epsilon}_t \sim \mathcal{N}(\mathbf{0}, \text{Q})
\end{equation}

where $f$ represents nonlinear dynamics, and $\boldsymbol{\epsilon}_t$ captures process noise modeled as zero-mean Gaussian distribution with covariance $\text{Q}$.

Given an initial state distribution $\mathbf{x}_0 \sim \mathcal{N}(\bar{\mathbf{x}}_0, \text{Q})$, the objective is to find a control sequence \( \mathbf{U} = \{ \mathbf{u}_0, \dots, \mathbf{u}_{T-1} \} \) that minimizes the expected cost over a finite horizon \( T \):
\vspace{-5pt}
\begin{equation}\label{eq:general_problem}
\begin{aligned}
    \min_{\mathbf{U}} \quad & J(\mathbf{U}) = \mathbb{E}_{u \sim \pi(\cdot|\mathbf{x}_t)} \left[ \phi_T(\mathbf{x}_T) + \sum_{t=0}^{T-1} c(\mathbf{x}_t, \mathbf{u}_t)  \right] \\
    \text{subject to} \quad & \mathbf{x}_{t+1} = f(\mathbf{x}_t, \mathbf{u}_t) + \boldsymbol{\epsilon}_t, \quad \boldsymbol{\epsilon}_t \sim \mathcal{N}(\mathbf{0}, \text{Q}), \\
    & \mathbf{x}_0 \sim \mathcal{N}(\bar{\mathbf{x}}_0, \text{Q}), \\
    & \mathbf{u}_t \in \mathcal{U}, \quad \mathbf{x}_t \in \mathcal{X}, \quad \forall t = 0, \dots, T-1
\end{aligned}
\end{equation}

where $c(\cdot, \cdot)$ is the running cost and  $\phi_T(\cdot)$ is the terminal cost. While the above formulation formulates the general setting, solving it requires systematic exploration. To this end, we introduce ~\ourmethodlong{} (\ourmethodto{}).

%% file: sections/approach.tex
In this section, we explain the core concepts of~\ourmethodto{}. It has two main stages. First, a set of candidate trajectories is generated by perturbing the nominal control sequence. Then, each trajectory is represented as a distribution by propagating its mean state and uncertainty ellipsoids through the system dynamics. This propagation captures both process noise and action selection. In the second stage, these trajectories are refined through a distributional separation step, where the Hellinger distance is used to measure the overlap between distributions.
Fig.~\ref{fig:method_figure} demonstrates how these distributions are getting separated through time. Minimizing this overlap encourages diversity in trajectories, allowing them to cover more unique areas in the configuration space while maintaining their dynamic feasibility. Alg.~\ref{alg:uae-to} shows the pseudocode of a single iteration of the optimization method. Finally, the resulting set of trajectories will be used as a nominal trajectory selection for the sampling-based MPC, which is explained in detail in Sec.~\ref {sec:uaempc}. We start with the formulation of trajectory distribution modeling.  

\SetKwFor{ParFor}{for}{do in parallel}{end for}
\begin{algorithm}[h] 
\caption{Uncertainty Guided Exploratory Trajectory Optimization (\ourmethodto{})}
\label{alg:uae-to}
\DontPrintSemicolon
\KwIn{Nominal sequence $\bar{\mathbf{U}}$, horizon $T$, number of trajectories $N$, rollouts per trajectory $M$, 
covariances $\Sigma_u, Q$, divergence metric $\mathcal{H}^2(\cdot\|\cdot)$,}
\KwOut{Optimized nominal control sequence $\bar{\mathbf{U}}$}

\ParFor{$i=1,\dots,N$}{
    $\mathbf{U}^{(i)} = \bar{\mathbf{U}} + \eta^{(i)}, \ \eta^{(i)} \sim \mathcal{N}(0,\Sigma_u)$ \;
    \For{$t=1,\dots,T$}{
        $\mathbf{\bar{x}}_{t+1}^{(i)} = f(\mathbf{\bar{x}}_t^{(i)}, \mathbf{u}_t^{(i)})$ \;
        $\Sigma_{t+1}^{(i)} = A_t^{(i)} \Sigma_t^{(i)} A_t^{(i)\top} + Q$ \;
        $\mathcal{T}^{(i)} = \mathcal{N}(\bar{x}^{(i)}, \Sigma^{(i)})$
    }
}
\textbf{Distributional Separation:}\;
\ParFor{$i=2,\dots,N$}{ 
  \For{$m=1,\dots,M$}{
    $\mathbf{U}^{(i,m)} = \mathbf{U}^{(i)} + \eta^{(i,m)},\quad \eta^{(i,m)} \sim \mathcal{N}(0,\Sigma_u)$\;
    build $\mathcal{T}^{(i,m)} = \mathcal{N}\!\big(\bar{x}_c^{(i,m)}, \Sigma_c^{(i,m)}\big)$\;
  }
    \For{$m=1,\dots,M$}{
        $h^{(i,m)} = \;
        \sum_{n \neq i}^{N}\mathcal{H}^2\!\left(
          \mathcal{N}(x_c^{(i,m)},\,\Sigma_c^{(i,m)}),\;
          \mathcal{N}(\bar{x}^{(n)},\,\Sigma^{(n)})
        \right)$
    }
  $m^\star \;=\; \arg\max_{m \in \{1,\dots,M\}}\, h^{(i,m)}$\;
  $\mathbf{U}^{(i)} \gets \mathbf{U}^{(i,m^\star)}$\;
}
\end{algorithm}

\subsection{Trajectory Distribution Modeling}

Given a nominal control sequence $\bar{\mathbf{U}} = \{ \bar{\mathbf{u}}_0, \dots, \bar{\mathbf{u}}_{T-1} \}$, we generate $N$ sampled control sequences by injecting Gaussian perturbations:
\begin{equation}
    \mathbf{U}^{(i)} = \bar{\mathbf{U}} + \boldsymbol{\eta}^{(i)}, 
    \quad \boldsymbol{\eta}^{(i)} \sim \mathcal{N}(\mathbf{0}, \Sigma_u)
\end{equation}

Each control sequence is propagated through the system dynamics to obtain a trajectory $\mathbf{X}^{(i)} = \{\mathbf{x}_0^{(i)}, \dots, \mathbf{x}_T^{(i)}\}$:
\begin{equation}
    \mathbf{x}_{t+1}^{(i)} = f(\mathbf{x}_t^{(i)}, \mathbf{u}_t^{(i)}) + \boldsymbol{\epsilon}_t^{(i)}, 
    \quad \boldsymbol{\epsilon}_t^{(i)} \sim \mathcal{N}(\mathbf{0}, Q)
\end{equation}

We approximate trajectory uncertainty by propagating the covariance forward using the linearized dynamics. With $A_t^{(i)} = \left.\frac{\partial f}{\partial \mathbf{x}}\right|_{\mathbf{x}_t^{(i)},\mathbf{u}_t^{(i)}}$, the covariance update is:
\begin{equation}
    \Sigma_{t+1}^{(i)} = A_t^{(i)} \Sigma_t^{(i)} A_t^{(i)\top} + Q
\end{equation}

Thus, each trajectory is modeled as a Gaussian distribution:
\begin{equation}
    \mathcal{T}^{(i)} = \mathcal{N}(\boldsymbol{\mu}^{(i)}, \boldsymbol{\Sigma}^{(i)})
\end{equation}
with mean $\boldsymbol{\mu}^{(i)} = [\mathbf{x}_0^{(i)\top}, \dots, \mathbf{x}_T^{(i)\top}]^\top$ and block-diagonal covariance $\boldsymbol{\Sigma}^{(i)} = \mathrm{blockdiag}(\Sigma_0^{(i)}, \dots, \Sigma_T^{(i)})$.

\subsection{Distributional Separation}

To promote diversity, we refine each trajectory by generating $M$ additional candidate rollouts:
\begin{equation}
    \mathbf{U}^{(i,m)} = \mathbf{U}^{(i)} + \eta^{(i,m)}, 
    \quad \eta^{(i,m)} \sim \mathcal{N}(\mathbf{0}, \Sigma_u)
\end{equation}
which yield trajectory distributions $\mathcal{T}^{(i,m)}$.

Each candidate is evaluated by measuring its dissimilarity to the other trajectories. We use the squared Hellinger distance to quantify distributional overlap because it is symmetric and bounded in $[0,1]$, unlike KL-type divergences, which can be unbounded \cite{ding2023empirical}.
\begin{equation}
\begin{aligned}
    &\mathcal{H}^2(\mathcal{T}^{(i,m)}, \mathcal{T}^{(j)}) 
    = 1 - \frac{|\Sigma^{(i,m)}|^{1/4} |\Sigma^{(j)}|^{1/4}}{\left| \bar{\Sigma}_{(i,m,j)}\right|^{1/2}} \\
    & \quad \times \exp\!\left(-\tfrac{1}{8} \, \Delta\mu_{(i,m,j)}^\top 
        \big(\bar{\Sigma}_{(i,m,j)}\big)^{-1} 
        \Delta\mu_{(i,m,j)} \right)
\end{aligned}
\end{equation}
where $\Delta\mu_{(i,m,j)} = \mu^{(i,m)} - \mu^{(j)}$, and $\bar{\Sigma}_{(i,m,j)} = \tfrac{1}{2}\big(\Sigma^{(i,m)} + \Sigma^{(j)}\big)$. The separation score of candidate $(i,m)$ is:
\begin{equation}
    h^{(i,m)} = \sum_{j \neq i} \mathcal{H}^2\!\left(\mathcal{T}^{(i,m)}, \mathcal{T}^{(j)}\right)
\end{equation}

For each trajectory $i$, we select the candidate with the largest separation score:
\begin{equation}
    m^\star = \arg\max_{m \in M} h^{(i,m)}, 
    \quad \mathbf{U}^{(i)} \gets \mathbf{U}^{(i,m^\star)}
\end{equation}

This separation procedure is repeated for $K$ iterations to enhance the diversity of the trajectory set.

%% file: sections/uae_mpc.tex
We now incorporate the proposed \ourmethodto{} into a sampling-based Model Predictive Controller. We use the Model Predictive Path Integral (MPPI) algorithm as the baseline, and augment it with the \ourmethodto{}.  

We start with the initialization of a nominal control sequence $\bar{\mathbf{U}} = \{ \bar{\mathbf{u}}_0, \dots, \bar{\mathbf{u}}_{T-1} \}$. It is either initialized as a zero control sequence or it is obtained from the previous iteration. After the initialization, we apply the \ourmethodto{} procedure described in Alg.~\ref{alg:uae-to} to the nominal trajectory to generate distributionally separated trajectories that remain diverse and explore a broad region of the state space.

\begin{algorithm}[h]
\DontPrintSemicolon
\KwIn{Nominal sequence $\bar{\mathbf{U}}$, horizon $T$, refinement size $N,M$, rollout size $L$, covariances $\Sigma_u, Q$, temperature $\lambda$}
\KwOut{Updated nominal control sequence $\bar{\mathbf{U}}$}


\textbf{Apply \ourmethodto{}($\bar{\mathbf{U}},T, N,M, \Sigma_u, Q) \quad \gets Alg.~\ref{alg:uae-to}$}\;
Evaluate $J(\mathbf{U}^{(i)})$ for all $i$\;
Select the best rollout: $i^\star = \arg\min_i J(\mathbf{U}^{(i)})$\;
Update nominal: $\bar{\mathbf{U}} \gets \mathbf{U}^{(i^\star)}$\;

\ParFor{$l=1,\dots,L$}{
    Sample $\mathbf{U}^{(l)} = \bar{\mathbf{U}} + \eta^{(l)}, \; \eta^{(l)} \sim \mathcal{N}(0,\Sigma_u)$\;
    Propagate $\mathbf{U}^{(l)}$ through dynamics $f$ to obtain $\mathbf{X}^{(l)}$\;
    Compute cost $J(\mathbf{U}^{(l)})$\;
}
Compute weights $w^{(l)} = \exp(-J(\mathbf{U}^{(l)})/\lambda)$\;
Update nominal: $\bar{\mathbf{U}} \gets \bar{\mathbf{U}} + \tfrac{\sum_l w^{(l)} \eta^{(l)}}{\sum_l w^{(l)}}$\;

\caption{Uncertainty-Guided Exploratory MPC (\ourmethodmpc{})}
\label{alg:uae-mpc}
\end{algorithm}
\vspace{-5pt}
After refinement, each trajectory $\mathbf{U}^{(i)}$ is evaluated using the cost function in Eq.~\ref{eq:general_problem}.
Instead of performing the importance-weighted update as in general sampling-based MPCs, \ourmethodmpc{} selects the best rollout as the new nominal sequence:
\begin{equation}
    i^\star = \arg\min_{i \in \{1,\dots,N\}} J(\mathbf{U}^{(i)}), 
    \quad \bar{\mathbf{U}} \gets \mathbf{U}^{(i^\star)}.
\end{equation}

This emphasizes that the nominal trajectory is directly related to the lowest-cost feasible rollout after diversity enforcement. After the new nominal control sequence. MPPI generates $L$ candidate control sequences by injecting Gaussian perturbations:
\begin{equation}
    \mathbf{U}^{(l)} = \bar{\mathbf{U}} + \boldsymbol{\eta}^{(l)}, 
    \quad \boldsymbol{\eta}^{(l)} \sim \mathcal{N}(0,\Sigma_u).
\end{equation}
Each sequence is propagated through the dynamics $f$ in Eq.~\ref{eq:system_dynamics_general} obtain trajectories $\mathbf{X}^{(l)}$. As in the solution of the MPPI~\cite{mppi}, the control sequence is updated as the cost-weighted average:
\begin{equation} \label{eq:update_rule}
    \bar{\mathbf{U}} \gets 
    \bar{\mathbf{U}} + \frac{\sum_{l=1}^L w^{(l)} \boldsymbol{\eta}^{(l)}}
    {\sum_{l=1}^L w^{(l)}}, 
    \quad 
    w^{(l)} = \exp\!\left(-\tfrac{1}{\lambda} J(\mathbf{U}^{(l)})\right).
\end{equation}
Since \ourmethodto{} only affects the nominal control sequence, we kept the MPPI update rule unchanged. Finally, after obtaining the MPPI solution, we apply the first control $\bar{\mathbf{u}}_0$ to the system, as in general MPC practice, and then shift the control sequence horizon. This procedure repeats until the stopping conditions are met.   

%% file: sections/experiments.tex
We evaluate the \ourmethodmpc{} algorithm by studying the following questions through experiments.

\begin{enumerate}
    \item Can the \ourmethodmpc{} converge to a solution when initialized from a poor warm start, including cases where the initial control sequence drives the system in a direction opposite to the goal? (Sec.~\ref{sec:trajectory_optimization})
    \item Can the \ourmethodmpc{} algorithm adapt and perform reliably in dynamic and complex environments in both simulation and real-world end-to-end navigation scenarios? (Secs.~\ref{sec:mpc_exp},~\ref{sec:real_word})
\end{enumerate}
\vspace{-10pt}
\subsection{Experimental Setup}

\subsubsection{Baselines} \label{sec:baselines}
To evaluate our approach, we compare against three baseline controllers of increasing complexity: Model Predictive Path Integral control (MPPI)~\cite{mppi}, MoG-MPPI ~\cite{wang2021variationalinferencempcusing}, and Stein Variational MPC (SV-MPC)~\cite{svmpc}. MPPI samples action sequences through a unimodal Gaussian distribution. MoG-MPPI extends this by introducing multiple Gaussian components to capture multimodality. Finally, SV-MPC adds a repulsion term among particles to encourage diversity. This progression of baselines demonstrates the incremental effects of adding multimodality and repulsion, and provides a benchmark for assessing our distributional separation in the configuration space. All methods use the same trajectory sampling budget ($2048$). We evaluate every method with a fixed number of trajectory samples to ensure a fair comparison.  MoG-MPPI and SV-MPC use $16$ particles, each with $128$ samples. We adopt the same particle settings but adjust the sample allocations proportionally to the number of iterations, ensuring that the total budget remains constant while accounting for iterative updates. All baselines are implemented based on~\cite{svmpc}.

\subsubsection{Cost Function} 

The cost function $J$ has two main components: The state cost $\mathcal{C}_{\text{state}}(x_t)$, which includes both obstacle $\mathcal{C}_{\text{obs}}(x_t)$ and distance-to-goal cost $\mathcal{C}_{\text{dist}}(x_t, x_{\text{goal}})$. On the other hand, the second term is the action regulator $\mathcal{C}_{\text{u}}(u_t)$, which smoothes the action to reduce jitter in the resulting trajectory. Corresponding weights are assigned to each component to give relative importance to each term. Therefore, the total cost is calculated over a time horizon $T$ as follows:

\begin{equation}
\begin{aligned}
\label{eq:cost_function_unified}
    J(\tau, \mathbf{U}) &= \phi(\mathbf{x}_T) 
    + \sum_{t=0}^{T-1} \left( \lambda_{\text{u}} \mathcal{C}_{\text{u}}(\mathbf{u}_t) 
    +\mathcal{C_{\text{state}}}(\mathbf{x}_t^{\tau}, \mathbf{x}_{\text{G}}) \right),
\end{aligned}
\end{equation}
where a trajectory $\tau = F(\mathbf{x}_{\text{curr}}, \mathbf{U}) = \{\mathbf{x}_t, \mathbf{u}_t\}_{t=0}^{T}$ and the initial state is equal to the current state $\mathbf{x}_0 = \mathbf{x}_{\text{curr}}$. The terminal cost is $\phi(\mathbf{x}_T) = \mathcal{C_{\text{state}}}(\mathbf{x}_T^{\tau}, \mathbf{x}_{\text{G}})$.
We introduce an indicator variable $\delta_t$:
\vspace{-3pt}
\begin{equation}
\delta_t =
\begin{cases}
1, & \exists \, \mathbf{x}_i^\tau \in \{\mathbf{x}_i^\tau\}_{i=0}^{t-1} \ \text{s.t.}\ \mathbf{x}_i \text{ is in collision}, \\
0, & \text{otherwise},
\end{cases}
\end{equation}

and define the unified stage cost as:

\begin{equation}
\begin{aligned}
\mathcal{C_{\text{state}}}(\mathbf{x}_t^\tau, \mathbf{x}_{\text{G}}) &=
\delta_t \left( \lambda_{\text{obs}}\mathcal{C}_{\text{collided}} + \lambda_{\text{dist}}\mathcal{C}_{\text{dist}} \right) \\
&+ (1-\delta_t) \left( \lambda_{\text{obs}}\mathcal{C}_{\text{obs}}(\mathbf{x}_t^\tau) + \lambda_{\text{dist}}\|\mathbf{x}_t^\tau - \mathbf{x}_{\text{G}}\| \right)
\end{aligned}
\end{equation}
where the indicator variable acts as a switch, if a state is in a collision, it will turn the remaining part of the trajectory into a collision, and the distance to goal value of the collided state is used as the distance-to-goal cost for the remaining part of the trajectory. $\mathcal{C}_{\text{collided}} = 10^3$ is the maximum collision cost, and $\mathcal{C}_{\text{obs}}(\mathbf{x}_t^{\tau})$ calculates the cost of the robot footprint of the state based on the local costmap. The term $\mathcal{C}_{\text{dist}}$ denotes the distance-to-goal cost of the state where the collision occurred along a trajectory $\tau$. It is important to note that if any state in a trajectory $\tau$ reaches the goal, we stop the cost calculation. This means the trajectory cost is measured only up to the goal-reaching state and terminal cost is calculated as $\phi(\mathbf{x}_T) = \mathcal{C_{\text{state}}}(\mathbf{x}_g^{\tau}, \mathbf{x}_{\text{G}})$ where $\statevec_g$ is the first goal reaching state along a trajectory $\tau$.

\begin{table*}[t]
    \caption{\small Per-goal performance comparison in open and cluttered environments. 
    The average goal-reaching time is computed using only the successful trials. 
    The best results are shown in bold.}
    \resizebox{\textwidth}{!}{%
    \begin{tabular}{c|
        c c|c c|c c|c c|c c|c c}
    \toprule
    \multirow{3}{*}{\textbf{Method}} & 
    \multicolumn{2}{c|}{Goal $(6,0)$} &
    \multicolumn{2}{c|}{Goal $(6,-4)$} &
    \multicolumn{2}{c|}{Goal $(-6,-4)$} &
    \multicolumn{2}{c|}{Goal $(-6,0)$} &
    \multicolumn{2}{c|}{Goal $(-6,4)$} &
    \multicolumn{2}{c}{Goal $(6,4)$} \\
    \cmidrule{2-13}
    & Open & Cluttered & Open & Cluttered & Open & Cluttered &
      Open & Cluttered & Open & Cluttered & Open & Cluttered \\
    \cmidrule{2-13}
    & SR($\uparrow$)/Time($\downarrow$) & SR($\uparrow$)/Time($\downarrow$)
    & SR($\uparrow$)/Time($\downarrow$) & SR($\uparrow$)/Time($\downarrow$)
    & SR($\uparrow$)/Time($\downarrow$) & SR($\uparrow$)/Time($\downarrow$)
    & SR($\uparrow$)/Time($\downarrow$) & SR($\uparrow$)/Time($\downarrow$)
    & SR($\uparrow$)/Time($\downarrow$) & SR($\uparrow$)/Time($\downarrow$)
    & SR($\uparrow$)/Time($\downarrow$) & SR($\uparrow$)/Time($\downarrow$) \\
    \midrule
    MPPI     
        & \textbf{1.00} / 0.2 & \textbf{1.00} / 0.2
        & \textbf{1.00} / 0.3 & \textbf{1.00} / 0.4
        & \textbf{1.00} / 1.1 & 0.80 / 2.9
        & 0.00 / --            & 0.00 / --
        & \textbf{1.00} / 1.3 & 0.30 / 5.7
        & \textbf{1.00} / 0.3 & \textbf{1.00} / 0.4 \\
    \midrule
    MoG-MPPI 
        & \textbf{1.00} / 0.3 & \textbf{1.00} / 0.6
        & \textbf{1.00} / 0.7 & \textbf{1.00} / 1.2
        & \textbf{1.00} / 1.7 & \textbf{1.00} / 5.2
        & \textbf{1.00} / 5.3 & 0.60 / 8.2
        & \textbf{1.00} / 1.7 & \textbf{0.90} / 6.1
        & \textbf{1.00} / 0.6 & \textbf{1.00} / 1.3 \\
    \midrule
    SV-MPC   
        & \textbf{1.00} / 0.4 & \textbf{1.00} / 0.4
        & \textbf{1.00} / 0.6 & \textbf{1.00} / 1.3
        & \textbf{1.00} / 1.7 & \textbf{1.00} / 5.3
        & \textbf{1.00} / 4.3 & 0.60 / 8.6
        & \textbf{1.00} / 1.8 & \textbf{0.90} / 5.6
        & \textbf{1.00} / 0.7 & \textbf{1.00} / 1.4 \\
    \midrule
    \ourmethodmpc{} (Ours) 
        & \textbf{1.00} / \textbf{0.1} & \textbf{1.00} / \textbf{0.1}
        & \textbf{1.00} / \textbf{0.1} & \textbf{1.00} / \textbf{0.1}
        & \textbf{1.00} / \textbf{0.8} & \textbf{1.00} / \textbf{2.1}
        & \textbf{1.00} / \textbf{0.8} & \textbf{1.00} / \textbf{2.7}
        & \textbf{1.00} / \textbf{1.1} & \textbf{0.90} / \textbf{2.7}
        & \textbf{1.00} / \textbf{0.1} & \textbf{1.00} / \textbf{0.2} \\
    \bottomrule
    \end{tabular}%
    }
    \label{tab:to_results}
\end{table*}

\subsection{Trajectory Optimization} \label{sec:trajectory_optimization}
In this experiment, we evaluate the performance of the proposed approach compared to 
baseline methods in settings where the goal is reachable within the prediction horizon. This setting transforms the MPC formulation into a full, dynamically feasible trajectory optimization problem.
\begin{figure}[h]
    \centering
    \includegraphics[height=3cm ]{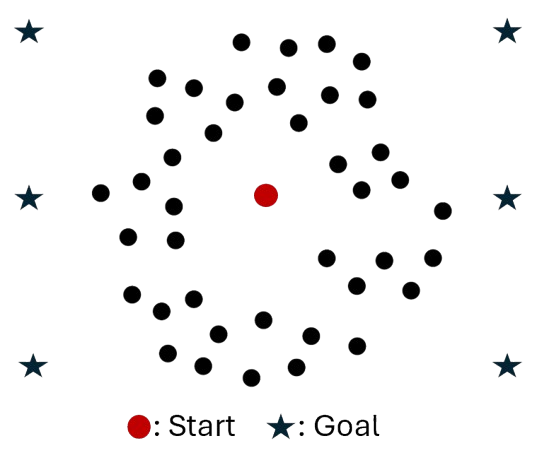}
    \caption{The cluttered environment with start and goal positions, where the black circles are arbitrarily placed obstacles.}
    \label{fig:env_to}
    \vspace{-2pt}
\end{figure}

We consider two 8x8 m environments: (i) an obstacle-free space and (ii) a cluttered environment. Fig.~\ref{fig:env_to} demonstrates
the cluttered environment
and the goal positions. The kinematic bicycle model is used for the dynamics. The start configuration is $[x,y, \theta] = [0,0,0]$ and the goal positions are selected in a circular manner to enforce nontrivial and challenging steering and velocity profiles. 
The nominal trajectory is initialized as a straight-line trajectory along the $+x$ direction, with a linear velocity $v=0.5~m/s$, and a steering angle, $\delta=0$. The trajectory horizon is $T=4s$ with the time discretization, $\Delta t = 0.05s$. Method parameters are set as follows: the temperature value for MPPI iterations is $\alpha = 0.001$, and the epsilon value for SV-MPC is $\epsilon =1.0$. In this experiment, we use the cost function $J$ as in Eq.~\ref{eq:cost_function_unified} without the action regularization term. The weights for the state and obstacle cost are assigned as follows: $\lambda_{\text{obs}} = 50$, and $\lambda_{\text{dist}} = 10$.

\vspace{-5pt}
\begin{figure}[h]
    \centering
    \includegraphics[width=\linewidth, height=7cm]{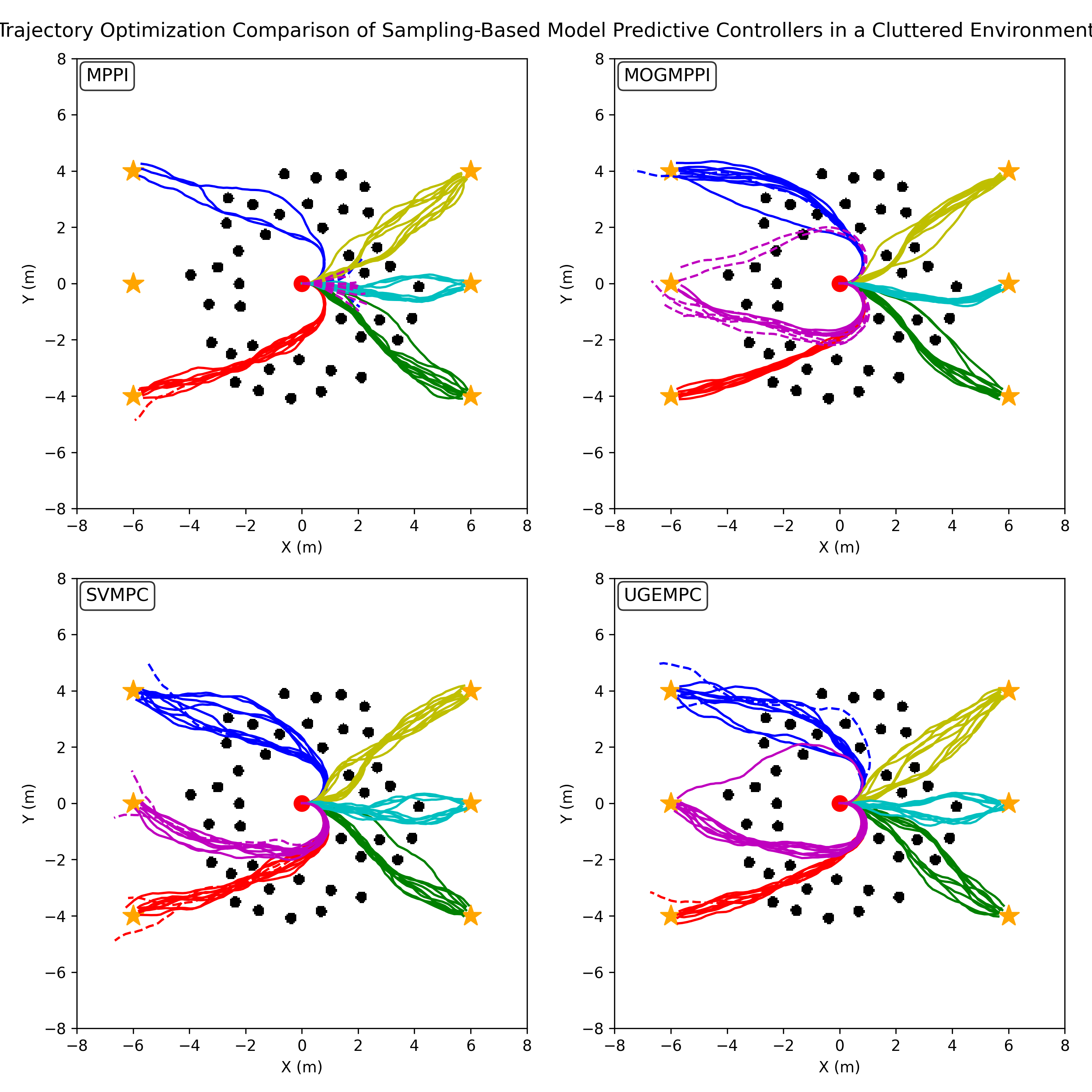}
    \caption{Trajectory optimization in a cluttered environment: Each start–goal pair is illustrated in a different color, with dashed lines marking unsuccessful runs. MPPI often struggles to find solutions, especially when the goal requires nontrivial velocity and steering profiles (e.g., full steering turns), due to its unimodal sampling distribution. Multimodal methods (MoG-MPPI, SV-MPC) achieve higher success rates but typically require longer times to converge. In contrast, \ourmethodmpc{} consistently generates systematically separated candidate trajectories that lead to faster convergence to a solution and provides non-colliding trajectories that improve robustness.}
    \label{fig:results_cluttered_to}
\end{figure}

We run 10 trials for each start-goal pair for every method. The maximum number of iterations is set to 100. If a method does not find a solution that reaches the goal within this limit, the trial is considered a failure. In Table~\ref {tab:to_results}, we report the average success rates and the average first goal reaching time. We highlight the best-performing methods in bold. The results indicate that, in the obstacle-free setting, every method can successfully generate trajectories that reach the goals. \ourmethodmpc{} outperforms all baselines in terms of average goal reaching time. This performance directly highlights that configuration space exploration helps find a candidate trajectory that reaches the goal more quickly than all baselines. Among the baselines, MPPI outperforms the other two multimodal methods in open-environment settings, where it selects a single mode, whereas the other baselines spend more time exploring multiple modes before converging.

Fig.~\ref{fig:results_cluttered_to} demonstrates the trajectories generated by each method in the cluttered environment. Each start and goal pair's trajectories are colored differently, where dashed lines show the corresponding unsuccessful runs.
In the cluttered environment scenario, \ourmethodmpc{} has achieved a higher success rate and lower average time compared to the baselines. On the other hand, in contrast to open space, now multimodal methods overperform the MPPI. Due to the unimodal structure of MPPI, it struggles to explore the environment, especially when the nominal trajectory is far from the optimal trajectory (The left-center goal in Fig.~\ref{fig:env_to}). At the same time, other baselines can shift their trajectories. Overall, these results demonstrate that our method,~\ourmethodmpc{}, achieves high success rates while comparably lower goal-reaching time, making it both efficient and robust in complex environments. 

\subsection{Model Predictive Control in Cluttered Environments}\label{sec:mpc_exp}

In this experiment, we analyzed the performance of our method against baselines in unknown cluttered environments. No prior map information was provided. We evaluated a point-to-point navigation task performance in settings where the vehicle only uses the local costmap derived from the lidar sensor footprint as a sensory input. To increase complexity, each environment (20x20m) was generated with concave polygonal obstacles, which introduced local traps in the environment. The start ($[2,2, \pi/2]$) and goal ($[18,18]$) positions are fixed for this experiment. The lidar sensor range was set to 10m, and each controller was run 10 times per environment to capture its behavior. Similarly to the previous trajectory optimization setting, we use trajectories with a $T=4s$ horizon with $0.05s$ as the time discretization. We set method-based hyperparameters as follows: $\alpha = 10^{-3}$ for the MPPI temperature parameters. The cost function Eq.~\ref{eq:cost_function_unified} without the action regularization term is used for this experiment, and the corresponding weights for obstacle and state cost are selected by $\lambda_{\text{obs}} = 10^3$, and $\lambda_{\text{dist}} = 10$.
\begin{figure}[h]
    \centering
    \includegraphics[width=\linewidth]{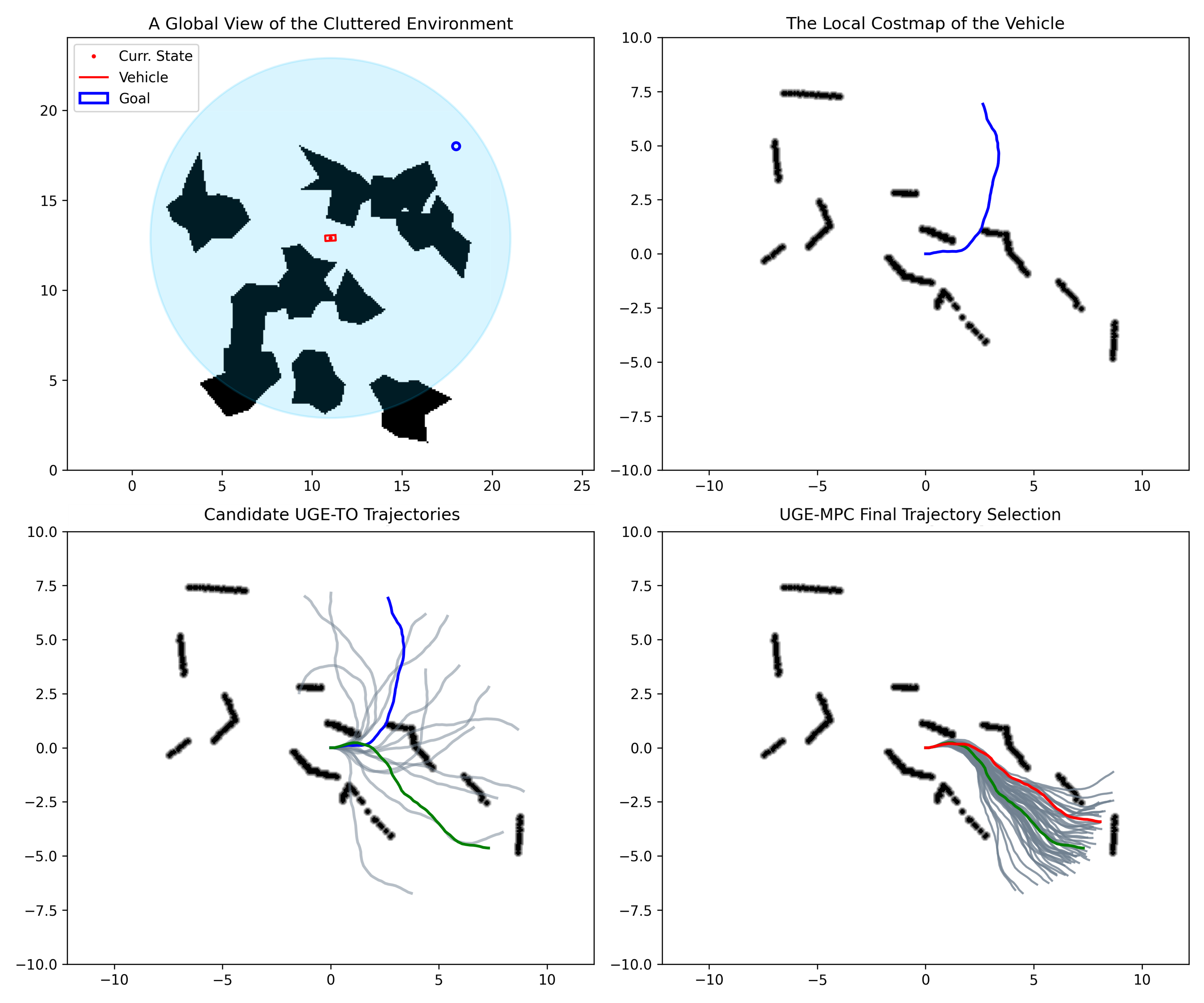}
    \caption{Visualization of Single iteration of \ourmethodmpc{}. \textbf{Top-left}: Global view of the environment with the current robot state (red), vehicle footprint (red polygon), and the goal region (blue circle). A semi-transparent blue circle shows the LiDAR range. \textbf{Top-right}: Local costmap with the nominal candidate trajectory (blue) expressed in the robot’s local frame before any optimization started. \textbf{Bottom-left}: \ourmethodto{} optimizes the candidate trajectories around the nominal trajectories to the trajectories in the figure by minimizing the overlap between their distribution. The lowest cost one is highlighted in green, while others are shown in gray, and the blue color is the nominal trajectory that comes from the previous time step solution. \textbf{Bottom-right}: Final step of \ourmethodmpc{} is selecting a final trajectory based on the update rule that is used in sampling-based MPC settings in Eq.~\ref{eq:update_rule}. After the nominal trajectory is selected with \ourmethodto{} iterations. We sample a new set of trajectories, and then the nominal trajectory is updated according to the rule. The new control sequence is highlighted in red.}
    \label{fig:uae_mpc}
    \vspace{-15pt}
\end{figure}

Fig.~\ref{fig:uae_mpc} shows an instance where \ourmethodto{} helps to generate diverse trajectories and shifts the nominal trajectory to one that avoids failure. Among these methods, \ourmethodmpc{} explored alternative configurations more effectively by explicitly encouraging distributional separation. That produces shorter and more direct trajectories. Table~\ref{tab:mpc_results} summarizes the average success rate and goal-reaching time across successful runs. 

\begin{table}[!h]
    \caption{\small Performance Comparison of Methods on Unknown Cluttered
    Environments. The average goal-reaching time is computed using only
    the successful trials.}
    \resizebox{\columnwidth}{!}{%
    \small
    \begin{tabular}{c|c|c}
    \toprule
    \multicolumn{1}{c|}{\textbf{Methods}}  & 
    \multicolumn{1}{c|}{\textbf{Success Rate ($\%$)($\uparrow$)}} &
    \multicolumn{1}{c}{\textbf{Average Goal-Reaching Time ($s$) ($\downarrow$)}} \\
    \midrule 
    MPPI & 0.54 & 17.3 \\
    \midrule
    MoG-MPPI & 0.60 & 24.5 \\
    \midrule
    SV-MPC & 0.64 & 25.6 \\
    \midrule
    \ourmethodmpc{} (Ours) & \textbf{0.88} & \textbf{21.0} \\
    \bottomrule
    \end{tabular}%
    }
    \label{tab:mpc_results}
\end{table}

\begin{figure}[t]
    \centering
    \includegraphics[width=0.95\linewidth,]{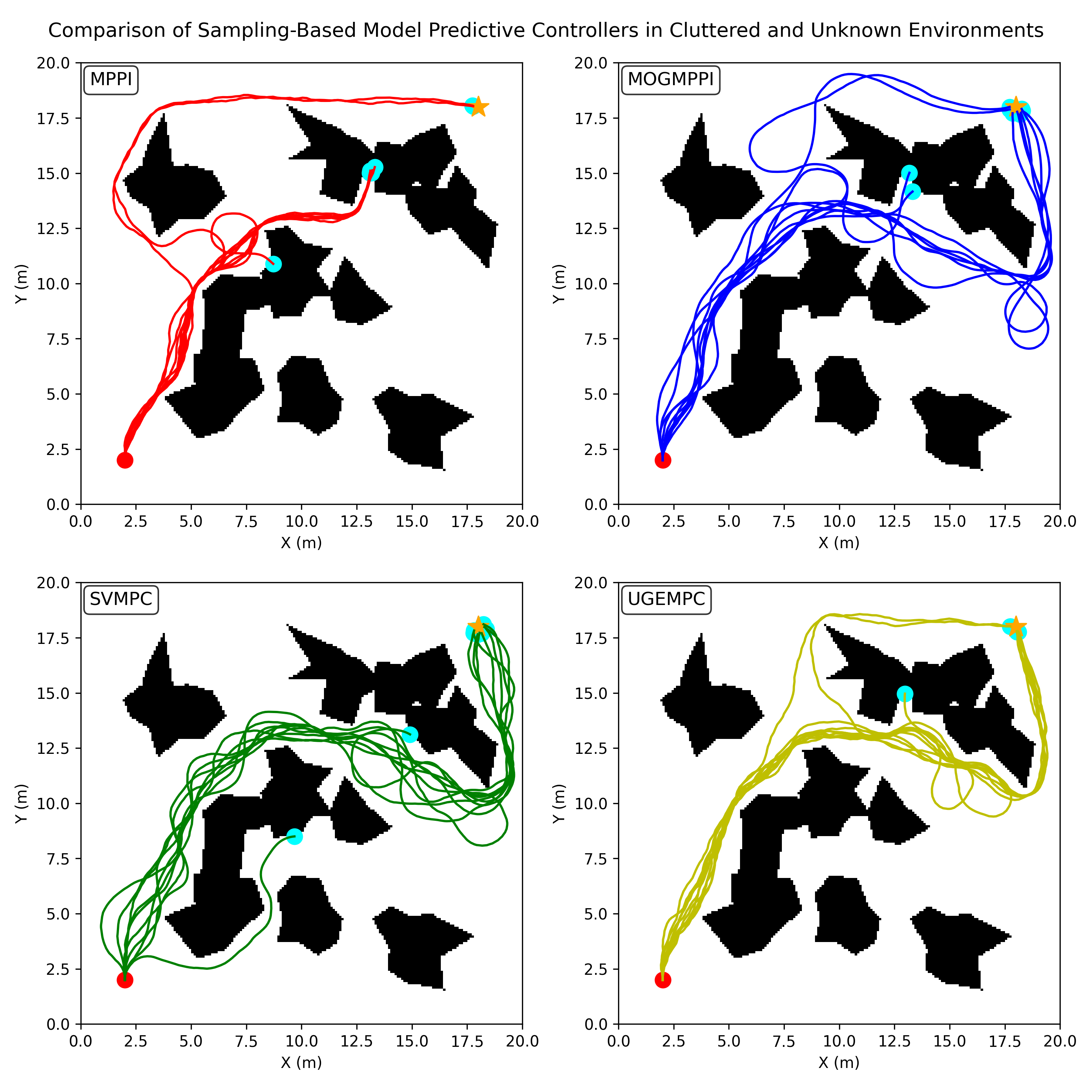}
    \caption{Performance Comparison in cluttered environments: It shows that MPPI, with its unimodal sampling, struggles to avoid local minima and may end up in one of those minima. On the other hand, MoG-MPPI, SVMPC, and~\ourmethodmpc{} achieve higher success rates by leveraging exploration.~\ourmethodmpc{} explores the configuration space more systematically and completes the task with shorter trajectories.}
    \label{fig:hard_env_mpc}
    \vspace{-10pt}
\end{figure}

Figure~\ref{fig:hard_env_mpc} illustrates a performance comparison in one of the five cluttered environments. It can be seen that MPPI, with its unimodal sampling distribution, often converges towards a single mode. In cases where the mode is a local trap, it is difficult for MPPI to escape from that local minimum. In contrast, the other baselines (MoG-MPPI and SVMPC) as well as our method, \ourmethodmpc{}, achieved higher success rates by leveraging exploration to escape such traps and adapt their nominal trajectories. 

\vspace{-4pt}
\subsection{Real-World Experiments} \label{sec:real_word}

\begin{figure}[h]
    \centering
    \includegraphics[height=4.0cm ]{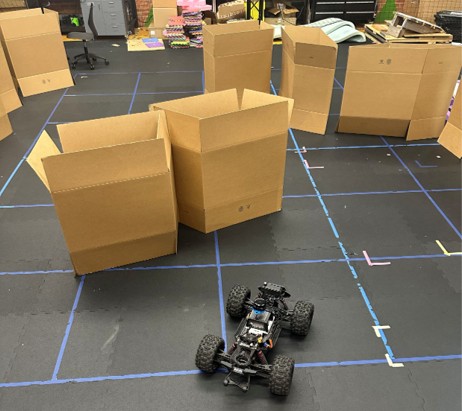}
    \caption{The $1/5$ scale truck platform and the cluttered environment setting}
    \label{fig:real_setup}
    \vspace{-10pt}
\end{figure}

We further evaluated our approach on a real-world point-to-point navigation task in a cluttered, unknown environment with fixed start and goal positions. We compared our method against two baselines, MPPI and log-MPPI, with five trials per method under a fixed sample budget of $N=512$. We allocated $126$ samples for the final MPPI iterations, using the remaining budget to generate $6$ candidate particles with $8$ iterations of~\ourmethodto{}. MPPI and log-MPPI covariances are both set to $\Sigma = [1.0, 10^{\circ}]$. Since log-MPPI uses a normal-log-normal distribution, it has higher exploration than MPPI. All controllers were run at 10 Hz to match the frequency of the onboard Slamtec R2 LiDAR. The $1/5$-scale vehicle and the environment are shown in Fig.~\ref{fig:real_setup}. Since SV-MPC and MoG-MPPI could not be executed consistently at the LiDAR frequency, they were excluded from the trials.

In all trials, MPPI became trapped in a local minimum and failed to reach the goal.  Both log-MPPI and~\ourmethodmpc{} consistently succeeded with average times $14.13s$ and $13.33s$, respectively. Under this setup, we did not observe a significant performance gap between log-MPPI and our approach. 

In summary, we conclude that the diversification of candidate nominal trajectories provided by our method is a viable approach to improve MPPI performance and avoid local minima.

%% file: sections/conclusion.tex
In this work, we introduced \ourmethodlong{} (\ourmethodto{}) and its integration into a sampling-based model predictive control framework,~\ourmethodmpc{}. We modeled trajectories as probability distributions and corresponding uncertainty ellipsoids. By distributionally separating them using a metric based on the Hellinger distance, our method enables systematic exploration of the configuration space beyond what existing action-space perturbation techniques can achieve. Through simulation and real-world experiments, we demonstrated that in most cases, \ourmethodmpc{} achieves higher success rates and faster convergence compared to MPPI, MoG-MPPI, and SV-MPC under the same sampling budget. In our future work, we aim to demonstrate \ourmethodmpc{} in much higher-dimensional systems, such as humanoid robots.